%% file: emnlp2021.tex
\pdfoutput=1

\documentclass[11pt]{article}

\usepackage{emnlp2021}

\usepackage{times}
\usepackage{latexsym}

\usepackage[T1]{fontenc}

\usepackage[utf8]{inputenc}

\usepackage{microtype}

\usepackage{amsmath}
\usepackage{amssymb}
\usepackage{url}
\usepackage{booktabs} 
\usepackage{graphicx}
\usepackage{epstopdf}
\usepackage{subfigure}
\usepackage{verbatim}
\usepackage{bm}
\usepackage{array}
\usepackage{multirow}
\usepackage{subfigure}
\usepackage{makecell}
\usepackage{xcolor}
\usepackage[most]{tcolorbox}

\usepackage[font={small}]{caption}

\newcommand{\abr}[1]{\textsc{#1}}
\newcommand{\nuanced}{\abr{nuanced}}

\title{NUANCED: Natural Utterance Annotation for Nuanced Conversation with Estimated Distributions}

\author{\textbf{Zhiyu Chen}\textsuperscript{1}\thanks{\hspace{5pt}Work done as a research intern at Facebook.}, \textbf{Honglei Liu}\textsuperscript{2}, \textbf{Hu Xu}\textsuperscript{2}, \textbf{Seungwhan Moon}\textsuperscript{2}, \textbf{Hao Zhou}\textsuperscript{2} and \textbf{Bing Liu}\textsuperscript{2} \\
  \textsuperscript{1}University of California, Santa Barbara \\
  \textsuperscript{2}Facebook \\
  {\tt zhiyuchen@cs.ucsb.edu}, \\ {\tt \{honglei,huxu,shanemoon,haozhoustat,bingl\}@fb.com} \\}

\begin{document}
\maketitle

\input{00-abstract.tex}
\input{01-introduction.tex}
\input{02-related.tex}
\input{03-dataset.tex}
\input{05-experiments.tex}
\input{06-conclusion.tex}

\section{Ethical Considerations}

For our data annotation, our annotators were hired as full-time employees through a leading annotation services vendor, and were paid in accordance with a fair wage rate.

\section*{Acknowledgment}
We thank Becka Silvert, Gerald Demeunynck, and Linnea Ross for helping with
the data annotation process. We thank the anonymous reviewers for their thoughtful comments.

\bibliography{emnlp2021}
\bibliographystyle{acl_natbib}

\input{07-appendix}

\end{document}

%% file: 00-abstract.tex
\begin{abstract}
Existing conversational systems are mostly agent-centric, which assumes the user utterances will closely follow the system ontology. However, in real-world scenarios, 
it is highly desirable that users can speak freely and naturally. 
In this work, we attempt to build a user-centric dialogue system for conversational recommendation.
As there is no clean mapping for a user's free form utterance to an ontology, we first model the user preferences as estimated distributions over the system ontology and map the user's utterances to such distributions.
Learning such a mapping poses new challenges on reasoning over various types of knowledge, ranging from factoid knowledge, commonsense knowledge to the users' own situations.
To this end, we build a new dataset named \nuanced{} that focuses on such realistic settings, with 5.1k dialogues, 26k turns of high-quality user responses. 
We conduct experiments, showing both the usefulness and challenges of our problem setting.
We believe \nuanced{} can serve as a valuable resource to push existing research from the agent-centric system to the user-centric system.
The dataset is publicly available\footnote{\url{https://github.com/facebookresearch/nuanced}}. 
\end{abstract}

%% file: 01-introduction.tex
\section{Introduction}
\begin{figure}[ht]
\centering
\includegraphics[width=0.48\textwidth]{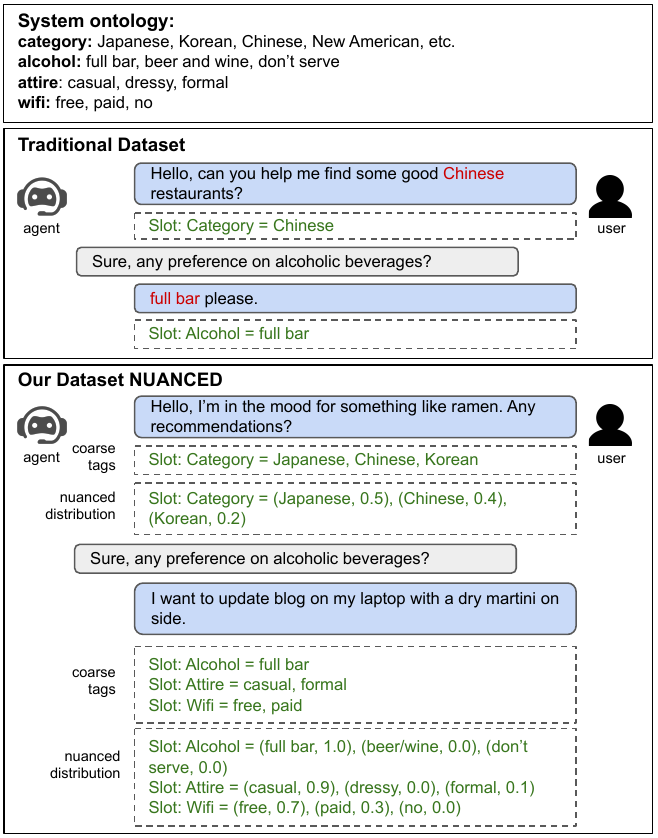}
\caption{Examples of 
traditional dataset and \nuanced{}: 
In \nuanced{}, we model the user preferences as distributions over the ontology to allow mapping of entities unknown to multiple values and slots for efficient conversation.} 
\label{fig:eg0}
\end{figure}


Conversational artificial intelligence is one of the long-standing research problems in natural language processing, such as task-oriented dialogue~\cite{DBLP:conf/icml/WenMBY17, DBLP:conf/emnlp/BudzianowskiWTC18, DBLP:journals/corr/abs-2005-00796}, conversational recommendation~\cite{DBLP:conf/sigir/SunZ18, DBLP:conf/cikm/ZhangCA0C18} and chi-chat \cite{adiwardana2020towards,roller2020recipes} etc. However, most existing systems are \textit{agent-centric}. Such systems require the users to \textit{unnaturally} adapt to and even have a learning curve on the system ontology, which is largely unknown to the users (such as the sample instructions for most smart speakers). Figure~\ref{fig:eg0} shows a dialogue snippet commonly found in traditional datasets: the user is \textit{expected} to closely follow the system ontology with the exact ontology terms, or at most with minor variations like synonyms. 

In the real-world use cases, such formulation may easily results in information loss, or breaks a conversation if the user speaks anything out of the system ontology; 
In this work, we argue that a smart agent can ideally be more \textit{user-centric}, by allowing users to speak freely without restrictions. The system is expected to uncover the connection between the freestyle user utterance and one or more slots and values by the system ontology. 

To build a \textit{user-centric} dialogue system, we propose to model the mapping from the free form user utterances to the system ontology as probability distributions to capture fine-grained user preferences. To learn the distributions, we construct a new dataset, named \nuanced{} (\textbf{N}atural \textbf{U}tterance \textbf{A}nnotation for \textbf{N}uanced \textbf{C}onversation with \textbf{E}stimated \textbf{D}istributions). \nuanced{} targets conversational recommendation because such type of dialogue system encourages more modeling of soft matching and implicit reasoning for user preference.
We employ professional linguists to annotate the dataset, and end up with 5.1k dialogues and 26k turns of high-quality user utterances. Our dataset captures a wide range of phenomena naturally occurring in realistic user utterances, including specified factoid knowledge, commonsense knowledge and users' own situations. We conduct comprehensive experiments and analyses to demonstrate the challenges. We hope \nuanced{} can serve as a valuable resource to bridge the gap between current researches and real-world applications. 


%% file: 02-related.tex
\section{Related Work}




\textit{Task-oriented dialogue systems} are typically divided into several sub modules, including user intent detection~\cite{DBLP:conf/interspeech/LiuL16, DBLP:conf/naacl/GangadharaiahN19}, dialogue state tracking~\cite{DBLP:conf/asru/RastogiHH17, DBLP:conf/sigdial/HeckNLGLMG20}, dialogue policy learning~\cite{DBLP:conf/emnlp/PengLLGCLW17, DBLP:conf/acl/SuGMRUVWY16}, and response generation~\cite{DBLP:conf/inlg/DusekNR18, DBLP:conf/emnlp/WenGMSVY15}. More recent approaches begin to build unified models that bring the pipeline together~\cite{DBLP:conf/acl/ChenCQYW19, DBLP:journals/corr/abs-2005-00796}. \textit{Conversational recommendation} focus on combining the recommendation system with online conversation to capture user preference~\cite{DBLP:conf/recsys/FuXZZ20, DBLP:conf/sigir/SunZ18, DBLP:conf/cikm/ZhangCA0C18}. Previous works mostly focus on learning the agent side policy to ask the right questions and make accurate recommendations, such as~\cite{DBLP:journals/corr/abs-2006-00184, DBLP:conf/kdd/LeiZ0MWCC20, DBLP:journals/corr/abs-2005-12979, DBLP:conf/recsys/PenhaH20}. Chit-Chat~\cite{adiwardana2020towards,roller2020recipes} is the most free form dialogue but almost with no knowledge grounding or state tracking. Both existing task-oriented, conversational recommendation systems have a pre-defined system ontology as a representation connected to the back-end database. The ontology defines all entity attributes as slots and the option values for each slot. In existing datasets, such as the DSTC challenges~\cite{DBLP:journals/aim/WilliamsHRTBR14}, Multi-WOZ~\cite{DBLP:conf/emnlp/BudzianowskiWTC18}, MGConvRex~\cite{DBLP:journals/corr/abs-2006-00184}, etc, the utterances from the users mostly closely follow the system ontology. While in task-oriented dialogue systems, parsing the user utterances into dialogue states is more on hard matching, in conversational recommendation systems soft matching is more encouraged since the user preferences are more salient and diverse in this type of conversations.

%% file: 03-dataset.tex
\section{The \nuanced{} Dataset}


\subsection{User Preference Modeling}
\label{formulation}
Given a system ontology, denote the set of all slots as $\{S_i\}$, with the option values for each slot as $\{V^j_i\}$. Denote the current user utterance as $T$ and dialogue context (of past turns) as $C$. 
We model the user preference as a distribution over each slot-value, namely \textit{preference distribution}: 
\begin{equation}
\begin{aligned}
    &P^j_i = P(V^j_i | T, C).
\end{aligned}
\end{equation}
Note that we expect the representation to be general, expandable, and to hold the fewest assumptions,
i.e., there is no assumption on the dependency among slot-values, nor the completeness of the value set. Therefore we model the distribution as a Bernoulli distribution over each slot-value. Intuitively, $P^j_i$ represents the probability that the user chooses an item with attributes $V^j_i$, under the observed condition of the dialogue up to the current turn. Note that the preference distributions may differ among individuals which causes variances, 
In this work, we aim to aggregate estimated distributions from large-scale data collected from multiple workers as ``commonsense'' distributions.
We leave modeling user-specific distributions to future work.

\subsection{Dataset Construction}
\label{construct}
We first simulate the dialogue flow with the preference distributions, then we ask the annotators to compose utterances that imply the distribution. 
\subsubsection{Dialogue Simulator}
We follow the approach from the MGConvRex dataset~\cite{DBLP:journals/corr/abs-2006-00184} to build the user visiting histories from real-world data. 
For each user with its visiting history as a list of restaurants with slot-values, we sample a subset of the history and aggregate to get a value distribution for each slot. For example, in the list of restaurants of a user’s visiting history, we sampled two restaurants, restaurant 1 and restaurant 2. Restaurant 1 has the slot-values of Alcohol = full\_bar, Restaurant 2 has the slot-value of Alcohol = beer\_and\_wine. Then the aggregated distributions is Alcohol = (full\_bar, 0.5), (beer\_and\_wine, 0.5), (no\_serve, 0.0). As generally, for the same user, the attributes of its visited restaurants tends to follow certain trends. Therefore the aggregated distributions created this way can be more natural. 
 Using the sampled distribution as the ground truth distribution, we simulate the dialogue skeletons of the following scenarios:
1) Straight dialogue flow: the system asks each slot, followed by the user response filled with preference distributions;
2) User updating preference: the user updates the preference distributions in a previous turn; 
3) System yes/no questions: the system can choose to ask confirmation questions;
For each turn, we randomly select 1 to 3 slots, corresponding to the cases that the user utterances naturally imply multiple slot-values. 
The system turns are composed using templates.

\subsubsection{User Utterances Composition}

After simulating the dialogue skeletons,
we employ professional linguists to do the composition to ensure high quality. We provide two composing strategies: 
\noindent \textbf{Implicit Reasoning}: do not mention the slot-value terms explicitly. This is the focus of this work because we expect that users are unaware of the system ontology and to depict their requests naturally.
\noindent \textbf{Explicitly Mention}: use the slot-value terms (or synonyms), as a backup option when the first one is not applicable. 
We also emphasize the following aspects: 
1) Read the whole dialogue first to have an overall ``story'' in mind before composing each utterance to ensure consistency; 2) Try to compose utterances as diverse as possible; 3) Reject any cases with invalid or unnatural preference distributions. 
We provide learning sessions to linguists to ensure they all master the tasks. 

\subsection{Dataset Statistics and Analysis}
\label{data_stats}
With an average of 5.39 user turns per dialogue, we have 5,100 dialogues consisting of 25,757 user turns. The user utterances have an average length of 19.43 tokens. 84.7\% of the utterances are composed using \textit{implicit reasoning}; 6.5\% of the utterances explicitly mention the ontology terms, and the rest use mixed strategies. 
The train / valid / test split is 3,600 / 500 / 1,000 in the number of dialogues, and 18,182 / 2,529 / 5,046 in the number of user turns.
To evaluate
the quality of our dataset, we randomly sample 500 examples 
and ask the linguistics
whether a preference distribution is reasonable based on the corresponding utterance. We end up with a turn-level correctness rate of 90.2\%. 

\begin{table*}[htbp]
\small
\begin{center}
\resizebox{.96\textwidth}{!}{%
\begin{tabular}{lll}
\toprule
\textbf{Reasoning types} & \textbf{Example user utterances} & \textbf{Example preference distributions} \\
\midrule
{\makecell[l]{\textbf{Type \MakeUppercase{\romannumeral 1}} Factoid Knowledge\\(37.3\%)}} 
& \makecell[l]{I really want a \textcolor{blue}{G\&T or a Riesling}, \\ but I could also have a \textcolor{blue}{tonic water}.} & \makecell[l]{Slot: Alcohol = (full\_bar, 0.7), (beer\_and\_wine, 0.2), \\(don’t\_serve, 0.1)} \\ 
\midrule
{\makecell[l]{\textbf{Type \MakeUppercase{\romannumeral 2}} Commonsense knowledge \\or User Situations\\(43.8\%)}} & \makecell[l]{ \textcolor{orange}{five to ten dollars}, I don't want a \\\textcolor{orange}{place with people wearing ties}, you\\ know?} & \makecell[l]{Slot: Price = (cheap, 0.6), (affordable, 0.4), \\(moderately\_priced, 0.0), (expensive, 0.0) \\ Slot: Attire = (casual, 1.0), (dressy, 0.0), (formal, 0.0)} \\ 
\midrule
{\makecell[l]{\textbf{Type \MakeUppercase{\romannumeral 3}} Mixed Type \MakeUppercase{\romannumeral 1} \& \MakeUppercase{\romannumeral 2}\\(19.0\%)}} & \makecell[l]{I want to \textcolor{orange}{update blog on my laptop}, \\with a \textcolor{blue}{dry martini} on side.} & \makecell[l]{Slot: Wifi = (free, 0.7), (paid, 0.3), (no, 0.0) \\ Slot: Alcohol = (full\_bar, 1.0), (beer\_and\_wine, 0.0), \\(don’t\_serve, 0.0)} \\ 
\bottomrule
\end{tabular}
}
\caption{Examples of reasoning types. Type \MakeUppercase{\romannumeral 1} utterance: G\&T is only served in a full bar, while Riesling is a kind of wine and tonic water does not require alcohol options. Type \MakeUppercase{\romannumeral 2} utterance, `place without people wearing ties' indicates casual attire, and `five to ten dollars' indicates a price range of cheap or affordable. Type \MakeUppercase{\romannumeral 3} utterance, we need both kinds of reasonings. }
\label{table:reasoning_types}
\end{center}
\vspace{-3mm}
\end{table*}
Among the utterances involving implicit reasoning, we summarize 3 basic reasoning types. The examples are shown in Table~\ref{table:reasoning_types}. 
\noindent \textbf{Type \MakeUppercase{\romannumeral 1} (Factoid Knowledge)} is largely agreed on by people and is relatively stable. 
\noindent \textbf{Type \MakeUppercase{\romannumeral 2} (Commonsense Knowledge or User Situations)} may not be formally defined.
For example, a food item less than \$10 is considered cheap. In many cases, such knowledge needs to be inferred from a situation described by users.
\noindent \textbf{Type \MakeUppercase{\romannumeral 3} (Mix of Type \MakeUppercase{\romannumeral 1} and \MakeUppercase{\romannumeral 2})} may appear in a single utterance. 

\subsubsection{\nuanced-reduced}
We also provide a \textit{reduced} variant called \textbf{\nuanced-reduced}, by discretizing the distributions for preference into binary numbers. 
For all slot-values with a positive preference distribution\footnote{In practice we set a threshold of 10\%, because in the utterance composition stage a preference distribution lower than 10\% is generally considered ignorable.} we label them as 1.0, otherwise 0.0. 
This reduced variant does not have continuous probabilities to tell the nuanced differences but it still needs to map free form utterances to binary labels.
We conduct human evaluation by asking the annotators to decide which representation can better capture more fine-grained user preferences. 
As Table~\ref{table:version_compare} shows, \nuanced{} can better capture the nuanced information.
Note that in real applications, which version of the data to use may depend on requirements of the system, i.e., level of granularity for state representation. 

\begin{table}[htbp]
\small
\begin{center}
\resizebox{.4\textwidth}{!}{%
\begin{tabular}{ccc}
\toprule
\textbf{\nuanced{} win} & \textbf{\nuanced-reduced win} & \textbf{Tied}\\
\midrule
 54.7\% & 16.7\% & 28.6\% \\
\bottomrule
\end{tabular}
}
\caption{Human evaluation results of comparing two versions.}
\label{table:version_compare}
\end{center}
\vspace{-3mm}
\end{table}

%% file: 05-experiments.tex
\section{Experiments}
In this section, we conduct experiments on both versions of the datasets in ~\S\ref{NUANCED-reduced} and ~\S\ref{NUANCED}, respectively. 
\vspace{-6mm}
\subsection{\nuanced-reduced}
\label{NUANCED-reduced}
\subsubsection{Baselines}

\noindent \textbf{Exact match \& Random guess} We follow the preceding system query to make slot prediction; we then use an exact match to predict the slot-values; if no match is found, we apply a random guess. 

\noindent \textbf{BERT} ~\cite{DBLP:conf/naacl/DevlinCLT19}, 
The input is the concatenation of the slot name, current turn system question and user utterance, and the dialogue context of past turns. 
We add two types of prediction heads on the \texttt{[CLS]} token of BERT, one for slot prediction (whether the input slot is updated or not), and the other for the value prediction of each slot.
The loss is a combination of cross-entropy loss for slot prediction and mean squared error (MSE) loss for value prediction. During inference, we set up a threshold
to decide positive or negative predictions. 

\noindent \textbf{Transformer}~\cite{DBLP:conf/nips/VaswaniSPUJGKP17} We use the similar architecture as the BERT baseline but train the weights from scratch.

\noindent \textbf{Train-ConvRex} 
As MGConvRex dataset~\cite{DBLP:journals/corr/abs-2006-00184} has similar domain and ontology, we compare the BERT model trained on MGConvRex\footnote{We contacted the first author to obtain the dataset.} with that tested on \nuanced-reduced. 
We use this baseline to demonstrate the open challenges caused by users' free-form speaking.

We refer the readers to Appendix A for more details. 
For all baselines, we evaluate 
on the turn level slot prediction accuracy and joint accuracy. 

\subsubsection{Results for \nuanced-reduced}
As shown in Table~\ref{table:res_reduced}, the BERT model achieves the best performance as the external knowledge obtained from pre-training helps draw a better relevance between unrecognized entities from the user and entities from the agent. 
Train-ConvRex limits such mapping to system ontology,
indicating that existing dialogue datasets may limit what an agent can understand from users.
Lastly, by comparing with BERT without dialogue context (or past turns), we notice that context may help in learning better values but yields more noise for slot prediction. 

\begin{table}[htbp]
\small
\begin{center}
\resizebox{.48\textwidth}{!}{%
\begin{tabular}{lcc}
\toprule
\textbf{Baselines} & \textbf{Slot Accuracy (\%)} & \textbf{Joint Accuracy (\%)}\\
\midrule
Exact match \& Random guess & 48.83 & 4.84 \\
\midrule
Train-ConvRex & 38.70 & 4.02 \\
\midrule
Transformer & 74.14 & 21.52 \\
\midrule
BERT & 88.21 & 36.56 \\
\midrule
BERT w/o context & 88.78 & 34.99 \\
\bottomrule
\end{tabular}
}
\caption{Results on \nuanced-reduced. \textit{Slot Accuracy}: percentage of turns that all slots are correct; \textit{Joint Accuracy}: percentage of turns that all slots and values are correct. 
}
\label{table:res_reduced}
\end{center}
\vspace{-8mm}
\end{table}

\subsection{\nuanced{}}
\label{NUANCED}

\subsubsection{Baselines}
\noindent \textbf{Exact match \& Random guess} Similar to \nuanced-reduced, we assign a probability of 1.0 for matched values or random value otherwise.

\noindent \textbf{BERT, Transformer} Similar to \nuanced-reduced, we use MSE loss between the ground truth and the predicted distribution.

\noindent \textbf{Train-reduced-X} 
We train the model on \nuanced-reduced and test on \nuanced{} to see how data with binary states can infer states in the continuous space. 
We define a fixed number of X as the continuous number for all positive predictions. We experiment with X = 0.5 and 1.0.  

We keep the same evaluation for slot prediction. For value predictions, we evaluate the \textit{soft} average mean absolute error (MAE) between the ground truth distribution and the predictions.

\subsubsection{Results for \nuanced{}}
As in Table~\ref{table:res_real}, BERT reaches the best performance.
One interesting observation is that using the same model BERT, the slot prediction accuracy increases (from 88.21\% to 89.62\%) compared with training on the reduced version.
\nuanced{} helps to reduce the noise of sparse entities in context (past turns). This is probably because numbers in continuous space can draw more relevance among different entities. 
As we can see, Train-reduced-X has a much larger error. 
This indicates that simply adapting the results from the reduced state labels suffers from information loss, i.e., the nuanced differences in continuous distributions. 

\begin{table}[htbp]
\small
\begin{center}
\resizebox{.48\textwidth}{!}{%
\begin{tabular}{lcc}
\toprule
\textbf{Baselines} & \textbf{Slot Accuracy (\%)} & \textbf{\makecell{Correct slots \\mean MAE (1e-2)}}\\
\midrule
Exact match \& Random guess & 48.83 & 46.84 \\
\midrule
Train-reduced-1.0 & 88.21 & 40.72 \\
\midrule
Train-reduced-0.5 & 88.21 & 21.62 \\
\midrule
Transformer & 78.42 & 16.78 \\
\midrule
BERT & 89.62 & 14.20 \\
\midrule
BERT w/o context & 88.08 & 14.49 \\
\bottomrule
\end{tabular}
}
\caption{Evaluation results on \nuanced. \textit{Correct slots mean MAE} (lower the better): mean absolute error of predicted distribution for all correctly predicted updated slots; 
}
\label{table:res_real}
\end{center}
\vspace{-5mm}
\end{table}

\subsubsection{Analysis on Slots}
We study how the models perform on different kinds of turns, 
shown in Table~\ref{table:res_slot_num}.
Generally speaking, the turns with more slots are relatively harder to learn. 
The turns that update the preference in previous turns have the highest error, the preference distribution needs to be jointly inferred from the previous mention and the current turn. We also study the performance on each slot in Appendix B, and provide some case studies in Appendix C.

\begin{table}[htbp]
\small
\begin{center}
\resizebox{.48\textwidth}{!}{%
\begin{tabular}{lccccc}
\toprule
\textbf{Type of turn} & \textbf{all} & \textbf{1 slot} & \textbf{2 slots} & \textbf{3 slots} & \textbf{\makecell{updating \\ preferences}}\\
\midrule
Slot Accuracy(\%) & 89.62 & 96.67 & 78.91 & 67.65 & 90.61 \\
\midrule
Mean MAE(1e-2) & 14.12 & 14.06 & 13.55 & 14.20 & 15.63 \\
\bottomrule
\end{tabular}
}
\caption{Performance for different kinds of slots: \textit{all}: all kinds of turns; \textit{n slots}: turns that the user utterance jointly implies n slots; \textit{updating preferences}: turns that the user utterance updates the preference in previous turns. 
}
\label{table:res_slot_num}
\end{center}
\end{table}
\vspace{-5mm}
\subsubsection{Human Evaluation}
We further conduct a human evaluation on baseline models.
We first evaluate the model outputs of Transformer, BERT, and BERT w/o context, through pairwise comparison between the model predictions and the gold labels. 
The results on 200 samples are shown in Table~\ref{table:human_type}.
There is a large gap between the best-performing baseline and the gold reference, which indicates significant room for improvement for future research.
Further, we study the breakdown of predictions of BERT on 3 different types of reasoning.
As shown in Table~\ref{table:human_cat}, the type 1 utterances, that involve factoid knowledge, are relatively harder to learn. 
This is close to our intuition because factoid knowledge is huge (and keeps increasing) and the limited utterances in the dataset may not cover all of the knowledge.

\begin{table}[htbp]
\small
\begin{center}
\resizebox{.48\textwidth}{!}{%
\begin{tabular}{lccc}
\toprule
Methods & Model output win(\%) & Tied(\%) & Gold win(\%) \\
\midrule
Transformer & 10 & 9.5 & 80.5 \\
\midrule
Bert & 23.6 & 20.9 & 55.4 \\
\midrule
Bert w/o context & 19.5 & 9.6 & 70.9 \\
\bottomrule
\end{tabular}
}
\caption{Human evaluation results for the model predictions.
}
\label{table:human_type}
\end{center}
\end{table}
\begin{table}[htbp]
\small
\begin{center}
\resizebox{.42\textwidth}{!}{%
\begin{tabular}{lccc}
\toprule
Methods & Model output win(\%) & Tied(\%) & Gold win(\%) \\
\midrule
Type \MakeUppercase{\romannumeral 1} & 22.5 & 19.9 & 57.6 \\
\midrule
Type \MakeUppercase{\romannumeral 2} & 27.4 & 24.1 & 48.5 \\
\midrule
Type \MakeUppercase{\romannumeral 3} & 21.1 & 11.2 & 67.7 \\
\bottomrule
\end{tabular}
}
\caption{Human evaluation results for different reasoning types. Type \MakeUppercase{\romannumeral 1}: factoid knowledge; Type \MakeUppercase{\romannumeral 2}: commonsense knowledge or user situations; Type \MakeUppercase{\romannumeral 3}:  Mixed Type \MakeUppercase{\romannumeral 1} \& \MakeUppercase{\romannumeral 2}. 
}
\label{table:human_cat}
\end{center}
\end{table}



%% file: 06-conclusion.tex
\section{Conclusion and Open Problems}
Starting from our dataset, we believe the user-centric dialogue system is an open-ended problem and the following directions are worth pursuing: 
1) Preliminary experimental results indicate that to improve performance, it is promising to incorporate external domain texts into pre-trained models, for example, pre-training the model on domain corpora like restaurant descriptions and reviews. 
2) Although our dataset collects a large set of domain entity knowledge, we still cannot guarantee that it will cover the vast amount of unknown entities in the future. 
One idea is to incorporate a knowledge base (KB) in the form of data augmentation or modeling.
3) Through our large-scale dataset, although one can learn a general agreement of estimated distributions from the crowds, a more user-specific distribution would be more desirable. 
We believe providing a personalized solution is another proper next step to consider.

%% file: 07-appendix.tex
\newpage
\appendix

\section*{Appendix}

\subsection*{A Model Implementation and Training Details}
\begin{figure}[ht]
\centering
\includegraphics[width=0.48\textwidth]{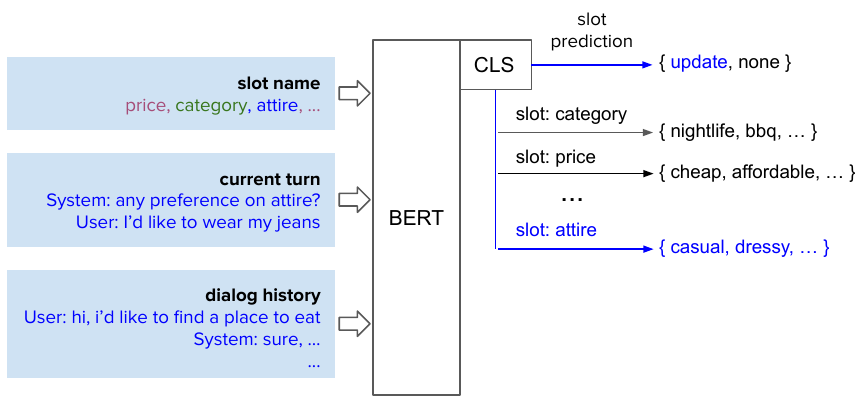}
\caption{Illustration of the BERT baseline} 
\label{fig:model}
\end{figure}

Figure~\ref{fig:model} presents the architecture of the BERT baseline. For each turn, we concatenate each slot with the current turn and the dialogue context as the input. On the \texttt{[CLS]} output, we add one head for slot prediction as binary classification, i.e., whether the input slot is updated in the current turn. For each slot, we add a specific head for value prediction. We use cross entropy loss for slot prediction, and mean squared loss for value distribution prediction. The overall loss is a weighted combination of the two losses. We set the weight for value prediction as 20.0. The threshold for value prediction in \nuanced-reduced is set as 0.5. We use BERT-base uncased model from the official release\footnote{https://github.com/google-research/bert} with 110M parameters; The learning rate is set as 3e-5, batch size as 32. We take the results based on the performance on validation set. For \nuanced-reduced, the training takes around 25,000 gradient steps; For \nuanced, the training takes around 40,000 steps. For the transformer model, to achieve best performance we use 6 layers and hidden size 300. All training is done on a single NVIDIA TESLA M40 card with 11G memory. 

Note that for the slot ``food category", some values are commonly observed in the dataset such as ``American food", ``nightlife", while some others are less frequently such as "Thai". During training we employ up-sampling for the less frequent ones.

In the construction of \nuanced, we sample a subset of the user history and aggregate to get the ground truth preference distributions.
Because the number of viable values of each slot is different, for those slots with relatively more values the distribution generally presents `long tail', we only take the top 3 value distributions for each slot. 
Correspondingly, during the model evaluation, we also take the top 3 predicted value distributions to calculate the MAE. 

\subsection*{B Analysis on Slots}
We also study how the model performs on each slot in the domain, shown in Table~\ref{table:res_slot_type}. 
Generally, slots that may involve more factoid knowledge or more choices of values are harder to learn, such as \texttt{food category}, \texttt{parking}. These may require learning long-tailed knowledge from external data.

\begin{table}[htbp]
\small
\begin{center}
\resizebox{.48\textwidth}{!}{%
\begin{tabular}{lcccc}
\toprule
\textbf{Slot} & \textbf{food category} & \textbf{price} & \textbf{parking} & \textbf{noise}\\
\midrule
Mean MAE(1e-2) & 15.48 & 15.29 & 16.94 & 13.34 \\
\toprule
\toprule
\textbf{Slot} & \textbf{ambience} & \textbf{alcohol} & \textbf{wifi} & \textbf{attire}\\
\midrule
Mean MAE(1e-2) & 15.04 & 13.88 & 12.30 & 8.95 \\
\bottomrule
\end{tabular}
}
\caption{Performance for each slot of our dataset. 
}
\label{table:res_slot_type}
\end{center}
\end{table}

\subsection*{C Case Studies}
Table~\ref{table:case_studies} provides some case studies with ground truth and the BERT model predictions.

\begin{table*}[htbp]
\small
\begin{center}
\resizebox{.96\textwidth}{!}{%
\begin{tabular}{lll}
\toprule
\textbf{Dialogue Turns} & \textbf{\nuanced-reduced} & \textbf{\nuanced} \\
\midrule
\multirow{2}{*}{\makecell[l]{\textbf{Assistant}: any preference on attire? \\ \textbf{User}: I like shorts and a loose tee shirt \\ in this heat.}} 
& \makecell[l]{\\ \textbf{Gold labels}: \\ Attire ( casual= 1, dressy= 0, formal= 0 )} & \makecell[l]{\\ \textbf{Gold Distributions}: \\ Attire ( casual= 1.00, dressy= 0.00, formal= 0.00 )} \\ \cline{2-3} & \makecell[l]{\\ \textbf{BERT predictions}: \\ Attire ( casual= 1, dressy= 0, formal= 0 )} & \makecell[l]{\\ \textbf{BERT predictions}: \\ Attire ( casual= 0.99, dressy= 0.01, formal= 0 )} \\
\midrule
\multirow{2}{*}{\makecell[l]{\textbf{Assistant}: what type of food would you \\like? \\ \textbf{User}: Ribs would be perfect.}} 
& \makecell[l]{\\ \textbf{Gold labels}: \\ Category ( traditional\_american= 1.0, bbq= 1.0, \\nightlife= 0.0 )} & \makecell[l]{\\ \textbf{Gold Distributions}: \\ Category ( traditional\_american= 0.50, bbq= 0.50, \\nightlife= 0.00 )} \\ \cline{2-3} & \makecell[l]{\\ \textbf{BERT predictions}: \\ category ( traditional\_american= 1.0, nightlife= 1.0, \\new\_american= 0.0 )} & \makecell[l]{\\ \textbf{BERT predictions}: \\ Category ( traditional\_american= 0.20, nightlife= 0.08, \\new\_american= 0.09 )} \\
\midrule
\multirow{2}{*}{\makecell[l]{\textbf{Assistant}: any preference on alcohol? \\ \textbf{User}: I really want a G\&T or a Riesling, \\ but I could also have a tonic water.}} 
& \makecell[l]{\\ \textbf{Gold labels}: \\ alcohol ( full\_bar= 1.0, beer\_and\_wine= 1.0, \\ don’t\_serve= 1.0 )} & \makecell[l]{\\ \textbf{Gold Distributions}: \\ alcohol ( full\_bar= 0.78, beer\_and\_wine= 0.33, \\ don’t\_serve= 0.11 )} \\ \cline{2-3} & \makecell[l]{\\ \textbf{BERT predictions}: \\ alcohol ( full\_bar= 1.0, beer\_and\_wine= 1.0, \\ don’t\_serve= 1.0 )} & \makecell[l]{\\ \textbf{BERT predictions}: \\ alcohol ( full\_bar= 0.55, beer\_and\_wine= 0.47, \\ don’t\_serve= 0.09 )} \\
\midrule
\multirow{2}{*}{\makecell[l]{\textbf{Assistant}: what parking option would \\ you like? \\ \textbf{User}: I need something fuss-free and \\ out of the rain for my car, Also, I really \\ want a gin and tonic, but it's not a \\ complete deal-breaker if I can't have it. \\ \\ \\ (after some turns) \\ \\ \textbf{Assistant}: here’re the recommendations. \\ \textbf{User}: You know what, if it's going to be \\ a fancier place then I don't mind dealing \\ with more complicated parking after all.}} 

& \makecell[l]{\\ \textbf{Gold labels}: \\ parking ( garage= 1.0, valet= 0.0, validated= 0.0 ) \\ alcohol ( full\_bar= 1.0, beer\_and\_wine= 1.0, \\ don’t\_serve= 1.0 )} & \makecell[l]{\\ \textbf{Gold Distributions}: \\ parking ( garage= 0.86, valet= 0.00, validated= 0.00 ) \\ alcohol ( full\_bar= 0.93, beer\_and\_wine= 0.21, \\ don’t\_serve= 0.14 )} \\ 
\cline{2-3} & 
\makecell[l]{\\ \textbf{BERT predictions}: \\ parking ( garage= 1.0, valet= 1.0, lot= 1.0 ) \\ alcohol ( full\_bar= 1.0, beer\_and\_wine= 1.0, \\ don’t\_serve= 1.0 ) } & \makecell[l]{\\ \textbf{BERT predictions}: \\ parking ( garage= 0.78, valet= 0.41, lot= 0.34 ) \\ alcohol ( full\_bar= 0.79, beer\_and\_wine= 0.17, \\don’t\_serve= 0.12 )} \\ 
\cline{2-3} & \\ &
\makecell[l]{\\ \textbf{Gold labels}: \\ parking ( garage= 1.0, valet= 1.0, validated= 1.0 ) } & \makecell[l]{\\ \textbf{Gold Distributions}: \\ parking ( garage= 0.86, valet= 0.64, validated= 0.21 )} \\ 
\cline{2-3} & 
\makecell[l]{\\ \textbf{BERT predictions}: \\ parking ( garage= 1.0, lot= 1.0, validated= 1.0 )} & \makecell[l]{\\ \textbf{BERT predictions}: \\ parking ( garage= 0.67, valet= 0.48, lot= 0.40 )} \\
\bottomrule
\end{tabular}
}
\caption{Some case studies. the last example shows two turns in a dialogue and corresponding distributions for each turn. The user updates the preference in a later turn based on a previous turn.}
\label{table:case_studies}
\end{center}
\end{table*}